\documentclass[10pt, a4paper]{article}
\usepackage{lrec}
\usepackage{graphicx}
\usepackage{tabularx}

\usepackage{soul}
\usepackage{titlesec}
\titleformat{\section}{\normalfont\large\bf\center}{\thesection.}{1em}{}
\titleformat{\subsection}{\normalfont\SmallTitleFont\bf\raggedright}{\thesubsection.}{1em}{}
\titleformat{\subsubsection}{\normalfont\normalsize\bf\raggedright}{\thesubsubsection.}{1em}{}
\renewcommand\thesection{\arabic{section}}
\renewcommand\thesubsection{\thesection.\arabic{subsection}}
\renewcommand\thesubsubsection{\thesubsection.\arabic{subsubsection}}
\usepackage{balance}
\usepackage{epstopdf}
\usepackage{subfig}
\usepackage{hyperref}
\usepackage{xstring}
\usepackage[table]{xcolor}
\usepackage{xcolor}
\usepackage{tabularx}
\definecolor{mygreen}{HTML}{488214}
\usepackage{pgfplotstable}

\usepackage{booktabs}
\title{Annotation of Emotion Carriers in Personal Narratives}
\definecolor{red1}{hsb}{0, 0.25, 1}
\definecolor{red2}{hsb}{0, 0.5, 1}
\definecolor{red3}{hsb}{0, 0.75, 1}
\definecolor{red4}{hsb}{0, 1, 1}
\usepackage[utf8]{inputenc}
\usepackage[T1]{fontenc}
\name{Aniruddha Tammewar$^1$, Alessandra Cervone$^1$, Eva-Maria Messner$^2$, Giuseppe Riccardi$^1$}

\address{
$^1$Signals and Interactive Systems Lab, University of Trento\\
$^2$Clinical Psychology and Psychotherapy, University of Ulm\\
         \{aniruddha.tammewar, alessandra.cervone, giuseppe.riccardi\}@unitn.it, eva-maria.messner@uni-ulm.de}

\abstract{
We are interested in the problem of understanding personal narratives (PN) - spoken or written - recollections of facts, events, and thoughts from one's own experience. 
For PNs, we define emotion carriers as the speech or text segments that best explain the emotional state of the narrator.
Such segments may span from single to multiple words, containing for example verb or noun phrases. 
Advanced automatic understanding of PNs requires not only the prediction of the narrator's emotional state but also to identify which events (e.g. {\em the loss of a relative} or {\em the visit of grandpa}) or people (e.g. {\em the old group of high school mates}) carry the emotion manifested during the personal recollection.
This work proposes and evaluates an annotation model for identifying emotion carriers in spoken personal narratives.
Compared to other text genres such as news and microblogs, spoken PNs are particularly challenging because a narrative is usually unstructured, involving multiple sub-events and characters as well as thoughts and associated emotions perceived by the narrator. In this work, we experiment with annotating emotion carriers in speech transcriptions from the Ulm State-of-Mind in Speech (USoMS) corpus, a German PN dataset. We believe this resource could be used for experiments in the automatic extraction of emotion carriers from PN, a task that could provide further advancements in narrative understanding.
\\ \newline  }

\pgfplotsset{compat=1.14}
\begin{document}
\maketitleabstract

\section{Introduction}

Emotion recognition is a well-grounded research field in the Speech and Natural Language Processing community. People express their emotions directly or indirectly through their speech, writings, facial expressions or gestures. Given such profusion of signals for expressing emotions, there has been work on analyzing emotions using various modalities such as speech, text, visuals, biosignals, and combinations of these \cite{poria2016convolutional,kudiri2016human,soleymani2015analysis}. Additionally the medium used as well as the genre might also have an effect on how emotions are conveyed. In text-based communication, for example, there are various mediums through which people might express their emotions, such as social media, personal diaries, news articles, blogs. Basic emotions are usually classified into classes such as \textit{anger, disgust, fear, joy, sadness, surprise} using established emotion annotation schemes \cite{ekman1992argument,shaver1987emotion,oatley1987towards}.

While most research on emotion analysis focuses on emotion classification, including predicting the emotions of the writer as well as those of the reader of a text \cite{chang2015linguistic}, a few studies so far have focused on identifying what might have triggered that emotion \cite{sailunaz2018emotion}. In particular, the task has been framed as \textit{Emotion Cause Extraction} \cite{lee2010text} from news and microblogs, where the cause of an emotion is usually a single clause \cite{chen2010emotion} connected by a discourse relation to another clause that explicitly expresses a given emotion \cite{cheng2017emotion}, as in this example from \newcite{gui2016event}: ``$<cause>$ \textit{Talking about his honours,} $</cause>$ \textit{Mr. Zhu is so} $<emotion>$ \textit{proud} $</emotion>$.''

However, if we move to other text genres beyond news and microblogs, it might be more complicated to identify the causes of given emotions from the text. In this work, we focus on spoken personal narratives.
A Personal Narrative (PN) can be defined as a recollection of an event or connected sequence of events the narrator has been part of, as an active or passive participant. Examples of PNs include personal diaries, short notes, travelogues in digital form (speech and/or text). Compared to other text genres such as news and microblogs, spoken PNs are particularly challenging because the structure of the narrative could be complex, involving multiple sub-events and characters as well as thoughts and associated emotions perceived by the narrator. 


\begin{table*}[ht]
\centering
\setlength{\tabcolsep}{12pt}
\begin{tabularx}{\textwidth}{@{}XX@{}}
\toprule
\multicolumn{1}{c}{\textbf{German (original)}} & \multicolumn{1}{c}{\textbf{English (translated)}} \\ \midrule
Okay. Ähm also eine Situation, in der ich mich kompetent gefühlt hab, war, als ich meinen ähm \colorbox{red2}{\textcolor{white}{\textbf{Praktikumsplatz}}} bekommen hab und ähm ich in dem \colorbox{red1}{\textcolor{white}{\textbf{Praktikum}}} dann auch ähm Sachen \colorbox{red4}{\textcolor{white}{\textbf{selbstständig}}} machen durfte und auch die Rückmeldung bekommen hab von den Personen dort, dass das, was ich da so mach, dass das gut ist und dass ähm sie mit mir sehr, mit mit mir sehr zufrieden sind ähm. Und die Gefühle dabei waren natürlich irgendwie Glück, weil man ist davor unsicher, ob man das, was man da macht, ob das so gut ist und ob man das so schafft. Ähm und das hat eben sehr gut funktioniert. Also ich hab mich sehr zufrieden gefühlt, mit mir ähm im Reinen, mit mir glücklich, auch irgendwie so ein bisschen \colorbox{red3}{\textcolor{white}{\textbf{Bestätigung}}} darin bekommen, dass das, was ich mach, gut ist oder das, was ich auch jetzt als Studium gemacht hab, irgendwie passt. Ähm ähm so bisschen so \colorbox{red2}{\textcolor{white}{\textbf{positive}}}\colorbox{red3}{ \textcolor{white}{\textbf{Aufregung}}}, also man fühlt sich sehr sehr wach, erregt irgendwie , aber in einer positiven Art und Weise. ... & OK. Um, so a situation in which I felt competent, was when I got my um \colorbox{red2}{\textcolor{white}{\textbf{internship position}}} and er in the \colorbox{red1}{\textcolor{white}{\textbf{internship}}} then I was also allowed to do things \colorbox{red4}{\textcolor{white}{\textbf{\textbf{independently}}}} and also got the feedback from the people there, that what I am doing there, that it is good, and that they are very pleased with me, with me, um. And of course the feelings were kind of happiness, because you are not sure if you, what you are doing, if that is so good and if you can do it that way. Um, and that worked very well. So I felt very satisfied, with me uh, at peace with myself, happy, with me, somehow getting a bit of \colorbox{red3}{\textcolor{white}{\textbf{confirmation}}} that what I'm doing is good or what I'm doing now as a study have, somehow fits. Uhm umh so a bit so \colorbox{red2}{\textcolor{white}{\textbf{positive}}}\colorbox{red3}{ \textcolor{white}{\textbf{excitement}}}, so you feel very very awake, excited somehow, but in a positive way. ... \\ \bottomrule
\end{tabularx}
\caption{\textbf{PN Annotated Example} (Colored Figure). A part of a narrative showing text spans annotated by the four annotators. The intensity of the red color in the background represents the number of annotators who annotated the text-span (varying from lightest for one annotator to the darkest for four annotators). It can be seen that some spans are annotated by only one annotator while some others by multiple. In the text-span \textit{``positive Aufregung''}, for example, two annotators selected the entire span while another one selected only the second word \textit{``Aufregung''}. It can be noticed how the annotations contain both sentiment words, such as \textit{``positive Augregung''} (i.e. positive excitement), and content words, such as \textit{``Praktikum''} (i.e. internship).}
\label{tab:annotation_example}
\end{table*}

\newcite{Tammewar2019}, for example, observed that in different machine learning models (Support Vector Machine, Attention-based neural sequence tagger) trained to predict valence (emotional value associated with a stimulus) from spoken PNs, concepts beyond sentiment words (\textit{sad}, \textit{happy}) were found to be useful. These concepts included terms such as characters (e.g. \textit{grandfather}, \textit{a friend}), locations (e.g. \textit{swimming pool}) and events (e.g. \textit{high school exam}).
As the task of the models was to predict the emotional state of the narrator, the authors concluded that these concepts played the role of explaining and {\em carrying} the emotional state of the person. We call such concepts \textit{emotion carriers}. Table \ref{tab:annotation_example}, shows a small part of a PN annotated with emotion carriers.

Inspired by such evidence, in this work, we investigate the possibility of annotating emotion carriers in spoken PNs. This type of annotation could then be used to experiment with training automatic Emotion Carriers Extraction systems. Such models, together with emotion prediction models, could provide a deeper understanding of the emotional state of the narrator.
The task of emotion carriers extraction can be classified as a task of Automatic Narrative Understanding (ANU), which encompasses tasks that extract various information from narratives \cite{fu2019asking}.
Emotion carriers extraction, for example, could be a useful task for conversational mental healthcare applications.
Applications aimed at the mental well-being of the users, often collect personal narratives from users in the form of personal diaries. In this setup, the applications may benefit from an analysis that provides not only the user's emotion trends but also the emotion carriers. The conversational agent of the application could use this information to start a conversation with the user and elicit more information about them, which in turn can be provided to a therapist for personalized interventions.


Nevertheless, annotating emotion carriers in PNs is a challenging task, since the same term may or may not carry emotions in different contexts and thus the search space for identifying emotion carriers becomes huge.

In this work, we present our experiments with manual annotation of emotion carriers in spoken PNs in the German language from the Ulm State-of-Mind in Speech (USoMS) corpus \cite{schuller2018interspeech}. After a review of relevant existing tasks in Section \ref{sect:related_work}, we describe the USoMs corpus in Section \ref{sect:USOMS}. The challenge of the task of Emotion Carriers extraction, formalized as the extraction of text segments perceived to be crucial for predicting the narrator's emotional state, is described in Section \ref{sect:task_description}, while the details about the manual annotation experiments are presented in Section \ref{sect:annotation}. In Section \ref{sect:analysis} we analyze and evaluate the annotated data using inter-annotator agreement metrics and qualitative analysis to get insights about the complexity of the task. Finally, in Section \ref{sect:conclusions}, we draw on the conclusions of our work.

\section{Related Work}
\label{sect:related_work}

While the emotion carriers in personal narratives is a new topic in the field of narrative understanding, the most relevant research to our task has been conducted on the task of \textit{Emotion Cause Extraction (ECE)}.
The task of ECE focuses on finding the cause of emotion from the given text. According to \newcite{talmy2000toward}, the cause of an emotion should be an event itself. The cause-event refers to the immediate cause of the emotion, which can be the actual trigger event or the perception of the trigger event \cite{lee2010text}. The cause events are further categorized into two types: verbal events and nominal events \cite{lee2010emotion}. 

There are a few emotion cause corpora on formal texts such as news reports, frames from FrameNet \cite{tokuhisa2008emotion,ghazi2015detecting,lee2010text,gui2016event} and informal texts such as microblogs \cite{gui2014emotion,gao2015rule,cheng2017emotion}. Some of these are annotated manually, while some are created automatically. 

While \newcite{lee2010text} defined the task of ECE as the extraction of word-level emotion causes, \newcite{chen2010emotion} suggested that a clause would be a more appropriate choice of unit for the extraction of an emotion cause. There have been works trying to solve the problem using different methods: Rule-based \cite{neviarouskaya2013extracting,li2014text,gao2015rule,gao2015emotion,yada2017bootstrap}; Machine Learning based \cite{ghazi2015detecting,song2015detecting,gui2016event,gui2016emotion,xu2017ensemble} and Deep Learning based \cite{gui2017question,li2018co,yu2019multiple,xu2019extracting}.

While most of the previous works are focused on either the news domain or microblogs, our work focuses on personal narratives. Personal narratives are more complex than the other domains as they are typically longer and contain multiple sub-events. Moreover, each sub-event has attributes (such as characters, entities involved in the sub-event), the narrator's reactions and emotions expressed in the narrative. In such a complex sequence of sub-events, it is difficult to associate the emotion clause with the corresponding event. It was one of the main shortcomings in the work by \newcite{gui2016event}, they call it the problem of \textit{cascading events}. Even if we succeed in correctly extracting the cause clauses for each sub-event, it is not our final goal. Our goal is to extract the emotion carriers used for conveying the emotions manifested in the narrator after recounting the entire narrative/event. The emotions produced by sub-events may or may not represent the emotions at the level of the entire narrative and thus the carriers as well. For instance, a narrative could begin with a happy event that is told to build context but could end up in a sad situation. Note that although the microblogs could also be considered as a personal narrative, we are dealing with longer personal narratives.

The previous approaches make one important assumption that an emotion keyword is always present in the text, for which they have to find the cause\cite{gui2016event}. As described by \cite{cheng2017emotion}, they consider the task of ECE as a discourse relation between the cause-clause and the emotion-clause. However, personal narratives may or may not contain emotion keywords. They might contain multiple keywords as well. Particularly in this task, the dataset is provided with the sentiment (positive or negative) of the text, and our goal is to find the emotion carriers for that sentiment.

\section{USoMs Corpus}
\label{sect:USOMS}
Ulm State-of-Mind in Speech (USoMs) is a database of spoken PNs in German, along with the self-assessed valence and arousal scores. A part of the dataset was used and released in the Self-Assessed Affect Sub-challenge, a part of the Interspeech 2018 Computational Paralinguistics Challenge (ComParE) \cite{schuller2018interspeech}. The task was to predict the narrator's valence score provided a short speech fragment (8 seconds) of the narrative. 

The data consists of 100 speakers (students) (85 f, 15 m, age 18-36 years, mean 22.3 years, std. dev. 3.6 years). In the challenge, the data was divided into three sets \textit{training, development, and test}. In this study, we annotate the development and the test sets (66 participants' data). The students told two negative and two positive PNs, each with a duration of about 5 minutes. Before and after recording each narrative, they self-assessed valence (spanning from negative to positive) and arousal (spanning from sleepy to excited) using the affect grid \cite{russell2003core} on a 10-point Likert scale. The narratives were transcribed manually. The number of tokens in the narratives vary from 292 to 1536 (mean: 820; std: 208). We use these transcripts in our work to enrich the annotations with the emotion carriers.\footnote{We plan to organize a challenge and release the annotated data as a part of it}

Following prompts were used to elicit the narratives 1) Negative narrative: \textit{"Please remember a time in your life when you were facing a seemingly unsolvable problem and report as detailed as possible over the next five minutes"}. 2) Positive narrative:\textit{"Please report of a time in your life were you found a solution, where you felt powerful, happy, and content. Describe that story in-depth over the next five minutes".}

\section{The challenge of Emotion Carrier Extraction}
\label{sect:task_description}
While describing an emotional event, to convey the emotions narrators not only use explicit emotion words such as \textit{happy, sad, excited} but also other emotion carriers such as entities, persons, objects, places, sub-events related to the event. \cite{Tammewar2019}, in their work, found that these carriers play an important role in predicting the current mental state of the narrator. There could be mentions of many such carriers in the narrative, but this does not imply that all of them reflect/carry the narrator's emotions. We call the text spans that capture the annotator's emotion as \textit{emotion carriers}. 

Apart from the carriers themselves, other factors also influence their importance, such as a context (from the narrative), the position of the span in the narrative, or the frequency of the mentions. Consider the term \textit{grandfather}. The emotional value/valence associated with the term changes according to the provided context, it can carry emotion accordingly. Consider the phrases \textit{``my grandfather died''} and \textit{``my grandfather came to visit us after a long time''}. In the first case, it might carry a negative emotion while, in the second case, a positive emotion. Often, narratives are longer, and people mention some terms to build the context for the main event. Consider the example \textit{``That day, my grandfather had come to visit us after a long time... we all went to the beach, where a saw a dreadful accident. A boy was swept out to sea while walking on the outer banks with his mom.... since then, I'm afraid of the water bodies.''} Here, the term \textit{grandfather} is used just to ground the context, while the emotion-relevant linguistic carriers come later in the narrative, such as \textit{``swept out''}, \textit{``water bodies''}. This uncertainty makes the task of identification of emotion carrying terms, complex and subjective. 

\section{Annotation of Emotion Carriers}
\label{sect:annotation}
To investigate the possibility of identifying emotion carriers from personal narratives, we annotate the personal narratives from the USoMs Corpus explained in Section \ref{sect:USOMS}, with the text spans that carry the emotions of the narrator manifested during the recollection.

In our annotation scheme, even though it is more relevant to us, we do not provide the annotators with a pre-selected noun or verb phrases to choose from.We give them the freedom to select text segments they feel are most important for our task.We believe that the pre-selection of spans could build bias in annotators towards specific fragments while there could be other text-fragments which are more important emotion carriers. Also, being spoken narratives, the automated tools to extract the noun and verb phrases may produce errors, thus affecting the annotation quality.

In this section, we provide details of the annotation experiments, including the annotators, the annotation scheme, and a brief overview of the tool used for the annotation.
\subsection{Annotators}
\label{sect:annotators}
Each narrative is annotated by four annotators. All the annotators are native German speakers and hold a Bachelor's degree in Psychology. They have been specifically trained to perform the task. We refer to the four annotators \textit{`ann1', `ann2', `ann3'} and \textit{`ann4'}
\subsection{Annotation Guidelines}
\label{sect:guidelines_scheme}
The annotation task involves the selection of the emotion carrying text spans as perceived by the annotator. We provide annotators with the guidelines to follow while performing the task.

We ask them to select sequences of adjacent words (one or more) in the text that explain why the narrative is positive or negative for the narrator.\\ We are particularly interested in words that play an important role in the story, such as:
\begin{itemize}
    \item  People (e.g.`mother', `uncle John', `my best friend'); Locations (`university', `our old school'); Objects (e.g. `guitar', `my first computer'); Events (`exam', `swimming class', `prom night')
    \item A clause that can include a verb and nouns (e.g. `Mary broke my heart', `I lost my guitar', `I failed the admission exam')
\end{itemize}
They have to select a minimum of three such text spans.\\
We also provide them with the best practices to be followed:
\begin{itemize}
    \item We ask them to annotate the contentful words (`university', `mother') preferably over pronouns (`she', `her', `it')
    \item If the same term is present multiple times, they are asked to annotate the first instance of the same concept and to avoid repetition.
    \item To make sure if something needs to be added or removed from the list of selected fragments, the annotators are asked to make sure:
    \begin{itemize}
        \item If a person who has not read the narrative can understand why the event was positive or negative just by looking at the list of spans they have selected. If not, they have to check if something is missing.
        \item They are asked to ensure that there are no repetitions in the list and that there are no spans, which are not central to the narrative.
    \end{itemize}
    \item As the annotators already know if the narrative is positive or negative in general, we ask them to annotate the feelings (emotion words) only if they are more informative (e.g. `feeling of freedom') than simple positive/negative (e.g. `I was happy').
\end{itemize}

 \subsection{Annotation Tool}
 \label{sect:tool}
 We provide the annotators with a web-based tool to perform the annotations. The tool is mainly divided into two parts. In one part, we show them a personal narrative and the corresponding sentiment. The annotator can hover over the tokens and select text spans by clicking and dragging over the consecutive tokens. On the right-hand side, they can see the already selected spans and their rankings. They can change the ranking by simple drag and drop.

Table \ref{tab:annotation_example} shows an example of annotations for a part of a narrative. We observe that sometimes, annotators annotate text-segments representing a similar concept but are at different positions in the text. In the example, the terms \textit{Praktikumsplatz} and \textit{Praktikum} represent the same concept of \textit{internship} but two of the annotators followed the guidelines to select the first occurrence while the other annotator selected the second occurrence of the same concept.

\section{Analysis}
\label{sect:analysis}
In this section, we perform an analysis of the annotations performed on the USoMs corpus from different perspectives. We take a look at some statistics of the annotations then we evaluate the annotations by calculating inter-annotator agreements with different strategies and finally we discuss some important observations we made.
\def\colorModel{hsb}
\def\cca#1{\cellcolor[\colorModel]{0,#1,1}\ifdim #1pt>0.5pt\color{white}\fi{#1}}

\begin{table*}[t]
\centering
\subfloat[Exact match, position agnostic, token level (mean $F_{1}$: 0.252)]{
\begin{tabular}{ccccc}
 & ann1 & ann2 & ann3 & ann4 \\
ann1 & \cca{1} & \cca{0.344} & \cca{0.417} & \cca{0.125} \\
ann2 &  & \cca{1} & \cca{0.389} & \cca{0.106} \\
ann3 &  &  & \cca{1} & \cca{0.137} \\
ann4 &  &  &  & \cca{1}
\end{tabular}}
\hspace{7em}
\subfloat[Partial match with position, token level (mean $F_{1}$: 0.320)]{
\begin{tabular}{ccccc}
 & ann1 & ann2 & ann3 & ann4 \\
ann1 & \cca{1} & \cca{0.338} & \cca{0.42} & \cca{0.277} \\
ann2 &  & \cca{1} & \cca{0.381} & \cca{0.196} \\
ann3 &  &  & \cca{1} & \cca{0.308} \\
ann4 &  &  &  & \cca{1}
\end{tabular}}
\qquad
\subfloat[Partial match, position agnostic, token level (mean $F_{1}$: 0.399)]{
\begin{tabular}{ccccc}
 & ann1 & ann2 & ann3 & ann4 \\
ann1 & \cca{1} & \cca{0.397} & \cca{0.483} & \cca{0.402} \\
ann2 &  & \cca{1} & \cca{0.439} & \cca{0.264} \\
ann3 &  &  & \cca{1} & \cca{0.404} \\
ann4 &  &  &  & \cca{1}
\end{tabular}}
\hspace{7em}
\subfloat[Partial match; position agnostic; lemma level (mean $F_{1}$: 0.403)]{
\begin{tabular}{ccccc}
 & ann1 & ann2 & ann3 & ann4 \\
ann1 & \cca{1} & \cca{0.400} & \cca{0.490} & \cca{0.410} \\
ann2 &  & \cca{1} & \cca{0.440} & \cca{0.267} \\
ann3 &  &  & \cca{1} & \cca{0.413} \\
ann4 &  &  &  & \cca{1}
\end{tabular}}
\caption{Pairwise Inter-Annotator Agreement scores ($F_1$ measure) with respect to the different matching strategies. We vary the matching criteria for annotations based on three aspects 1) checking if the annotations are exactly same (exact match) or calculating the overlap between them (partial match) 2) positions of the annotations in the text is considered whiles matching (with position) or not (position agnostic) and 3) to calculate the overlap, the tokens in the annotations are matched (token level) or the lemmas of the tokens are matched (lemma level). Note that in the tables b, c and d we report the soft $F_1$ measure, since we are calculating partial match rather than exact match.}
\label{tab:results}
\end{table*}
\subsection{Statistics}
\label{sect:stats}
In this study, we analyze 239 narratives from 66 participants (the development and test sets from the ComParE challenge) that have been annotated by four annotators each. Note that for 66 participants the total number of narratives should be 264, but in the ComParE challenge, 25 files were removed because of issues like noise.

We observe that the number of annotations (text-spans) annotated by the annotators per narrative vary from 3 to 14 with an average of 4.6, also that all annotators follow the same pattern from this aspect. We also calculated the number of tokens present in the annotations. The numbers show that three of the annotators (\textit{ann1, ann2, ann3}), on average select a span of 1.5 tokens, while the fourth annotator (\textit{ann4}) selects three tokens (avg.) per annotation. Note that, for all the analysis, we use the spaCy toolkit\footnote{\href{https://spacy.io/}{https://spacy.io/}} for tokenization. We observe that many annotations contain punctuation marks, which are considered as separate tokens by spaCy. Thus, we perform the same calculations while ignoring the punctuation tokens. We find that the average number of tokens drops down to 1.1 for the first three annotators, while it drops down to 2.3 for the \textit{ann4}.

We also analyzed the distributions of POS tags, and as expected, found that the most common categories include  noun (35\%), adjective (30\%), verb (15\%), and adverb (7\%).

\subsection{Evaluation of Annotations}
In this section we measure the quality of the annotations in terms of inter-annotator agreement. In the first part, we define different metrics for evaluation. In the second part, we show the results using different metrics and strategies.
\subsubsection{Metrics}
Commonly used metrics for evaluating the agreement between annotators include variations of $\kappa$ coefficient such as Cohen's \cite{cohen1960coefficient} for two annotators, Fleiss' \cite{fleiss1971measuring} for multiple annotators. Unfortunately, calculations for $\kappa$ such as observed and chance agreements involve the knowledge of true negatives, which is not well defined for a text span selection task. (eg. in this study, it could mean the number of possible text spans that are not annotated). This makes $\kappa$ impractical as a measure of agreement for text spans annotation.

An alternative agreement measure that does not require the knowledge of true negatives for its calculations is Positive (Specific) Agreement \cite{fleiss1975measuring}(${ P  }_{ pos  }$ Eq. \ref{eq:ppos}). It has previously been shown to be useful in the evaluation of crowdsourced annotations tasks, similar to our's \cite{stepanov2018cross,chowdhury2014cross}. 
\begin{figure*}[t]
    \centering
    \includegraphics[width=\textwidth]{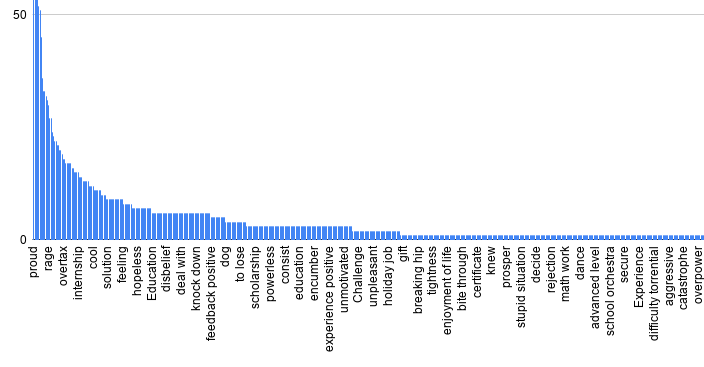}
    \caption{Sorted counts of emotion carriers (English translations of annotated German text spans) shared and agreed upon by the annotators. Notice the long tail of singletons. Due to space limitations, we perform binning of the carriers and show only a representative element from each bin.}
    \label{fig:histogram}
\end{figure*}

The Equation \ref{eq:ppos} defines the positive agreement in terms of true positives (TP), false positives (FP) and false negatives (FN). We can see that the knowledge of \textit{true negatives (TN)} is not required for the calculation of the positive agreement. Also, notice that the equation is similar to the widely used $F_1$-measure \cite{hripcsak2005agreement}. In our experiments, we calculate the positive agreement for each pair of annotators.
\begin{equation}
\label{eq:ppos}
{ P  }_{ pos  }   =   \frac{ 2\times TP  }{ 2\times TP + FP + FN  }
\end{equation}

Another problem we face in the task of text spans selection is the annotation of overlapping text fragments. Given the freedom on the lengths and positions of the text spans, two annotators might annotate different but overlapping text spans. The overlapping part could be an important part, thus the annotations should not be discarded completely. For instance, in Table \ref{tab:annotation_example}, \textit{`positive Aufregung'} and \textit{`Aufregung'}, both the spans contain the fragment \textit{`Aufregung'}, which is important to be considered. Thus, we report results on exact matches as well as partial matches, following the work by \cite{johansson-moschitti-2010-syntactic}. For the partial match, they calculate ``soft'' $F_1$-measure by calculating the coverage of the hypothesis spans.
The coverage of a span($s$) with respect to another span ($s'$) is calculated as defined in Equation \ref{eq:coverage}, with the help of the number of tokens common in the two spans. The operator$|.|$ counts the number of tokens.
\begin{equation}
\label{eq:coverage}
    c ( s , s ^ { \prime } ) = \frac { | s \cap s ^ { \prime } | } { | s | }
\end{equation}
Next, a span set coverage $C$ is defined for a set of spans $S$ with respect to another set of spans $S'$ using the Equation \ref{eq:set_coverage}.
\begin{equation}
\label{eq:set_coverage}
    C ( S , S ^ { \prime } ) = \sum _ { s _ { i } \in S } \sum _ { s _ { j } \in S ^ { \prime } } c ( s _ { i } , s _ { j } ^ { \prime } )
\end{equation}
In order to calculate the soft $F_1$-measure, first soft precision and soft recall are calculated according to Eq \ref{eq:softprec} and Eq \ref{eq:softrecall} respectively. Here $S_H$ and $S_R$ are hypothesis and reference spans respectively, and $|.|$ operator counts the number of spans.
\begin{equation}
\label{eq:softprec}   
    precision( S _ { R } , S _ { H } ) = \frac { C ( S _ { R } , S _ { H } ) } { | S _ { H } | }
\end{equation}
\begin{equation}
\label{eq:softrecall} 
    recall( S _ { R } , S _ { H } ) = \frac { C ( S _ { H } , S _ { R } ) } { | S _ { R } | }
\end{equation}
Finally the soft $F_1$-measure is calculated using the standard formula \ref{eq:soft_f}:
\begin{equation}
\label{eq:soft_f} 
    F_1  =   2 \times {\frac{precision \times recall}{precision + recall}}
\end{equation}
As the personal narratives are longer, often some terms are repetitive. In our task, the position of an annotation is not quite important compared to the content. We further try to loosen the criteria for matching by not considering the position of the text fragments. For instance, let us say a narrative contains mentions of \textit{`trip'} at multiple places, like \textit{`we went for a trip to India'} and \textit{`the trip was great'}. If two annotators intend to annotate the word \textit{`trip'}, they have multiple locations to choose from. While from the perspective of discourse, it would be interesting to see which position seems more appropriate, for our purpose of extraction of emotion carriers it is less important. Following the same intuition, we also try to match tokens having the same lemma. 

\subsubsection{Results}
Table \ref{tab:results} shows the evaluation results based on the various strategies of matching described above. The $F_1$-measure is calculated for all pairs of annotators. For each strategy, we also report the mean of pairwise scores. In the four tables from Table (a) to Table (d) we loosen the matching criteria, thus increasing the scores. We show the results starting from the most strict criteria of exact matching in the table (a), then in the table (b), we show results for partial matching, but the positions of the annotations are taken into consideration. The improvements are most significant in the case of \textit{ann4}, as we saw earlier in Section \ref{sect:stats} that \textit{ann4} usually annotates longer fragments than others. This shows that the \textit{ann4} annotates longer spans, but still contains the important part that other annotators annotate. Later in table (C), we remove the constraint of position, which results in improved scores, showing that even if the annotations by different annotators are different they often contain similar terms/carriers. This also shows that the annotators often ignore the instruction from the guidelines of selecting the first occurrence of the same term (Section \ref{sect:guidelines_scheme}). In the table (d), we further try to match more things by considering lemmas instead of tokens, which results in an increment.



\subsection{Discussion}
\label{sect:qualitative_analysis}
In this section, firstly we discuss the quality of emotion carriers annotations compared to other annotation efforts which used inter-annotator metrics similar to ours. Secondly, we explore two main questions in regards to emotion carriers annotations in our corpus, namely whether emotion carriers consists only of sentiment words and whether we can observe any positional relation between emotion carriers and disfluencies.

\paragraph{How good are the annotations?}
It is difficult to judge the quality of the annotations by looking at the inter-annotator agreement scores, which are not self-explanatory. If we compare our task with other previous tasks that used a similar metric, we can better understand the complexity of the task and judge the quality of the annotations.
\newcite{chowdhury2014cross} worked on the task of semantic annotations of the utterances from conversations. For example, one of the sub-task annotators had to perform was selecting a text span describing a hardware concept. For a particular concept like \textit{printer}, the annotators could select the spans \textit{`with the printer'}, \textit{`the printer'} or just \textit{`printer'}, all of which are correct. The problem they faced for the selection of span is similar to ours, but the complexity and subjectivity are low, as they work on shorter texts and the annotator has more idea about the concept to be selected. They use the same metric as ours to evaluate the inter-annotator agreement for the span selection. They achieve $F_1$ scores of 0.39 and 0.46 (for two different subsets of data) for the exact match, whereas 0.63 and 0.7 for the partial match (mean of pairwise agreements between three annotators). While our scores for exact and partial matches are 0.25 and 0.4, which we believe are reasonable given the subjectivity of the task and more number of annotators.

\paragraph{Are emotion carriers just sentiment words?}
As seen in the example from Table \ref{tab:annotation_example}, the annotations include sentiment words as well as content words.
In order to further study what is the actual distribution of sentiment words (\textit{angry, joy}) versus content words in emotion carriers, we analyze the annotation of sentiment words across the annotators.
For this, we calculate the sentiment polarity of each annotation using the textblob-de library \footnote{\href{https://textblob-de.readthedocs.io/en/latest}{https://textblob-de.readthedocs.io/en/latest}}, which makes use of the polarity scores of the words from senti-wordnet for German (with simple heuristics), similar to the English senti-wordnet \cite{esuli2006sentiwordnet}. We find that the trends of using sentiment carrying phrases vary across the annotators. The fraction of annotations carrying sentiment varies from 24\% to 56\% (\textit{ann1: 39\%; ann2: 24\%; ann3: 36\%; ann4: 56\%}) for the four annotators. For further analysis, we plan to categorize the annotations into categories inspired by the ones used in the \textit{Psychological Processes} categories of the LIWC dictionary \cite{pennebaker2015development}.

In addition to the distribution across the annotators, we further analyze if there is a trend in the counts of occurrences of sentiment and content words.
In Figure \ref{fig:histogram}, we show an interesting observation from the annotations. We plot the histogram of overlaps from partial-matches that we get while evaluating the inter-annotator agreements for all annotator-pairs, with respect to their counts of occurrences. Due to space limitations, we only show the representative annotations and not all the partial-matches. We can see that the overlaps contain both, emotion words such as \textit{proud, hopeless, disbelief} as well as content words like \textit{scholarship, education, dance}. We notice that there is a long tail of overlaps having only a single occurrence. This shows that a few terms are annotated frequently and agreed upon by annotators, while many terms are unique to specific narratives. As expected, we observe more content words than sentiment words in the tail, while it is surprising to see content words like \textit{internship, solution} appearing in the most frequent words.
\begin{figure}[ht]
    \centering
    \includegraphics[width=\columnwidth]{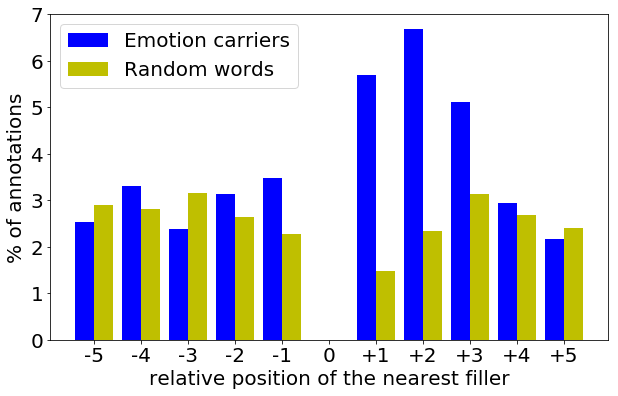}
    \caption{The relative positions of the closest fillers (a type of disfluency, e.g. ähm, äh, mhm) with respect to emotion carriers (in blue) and random words (in green) in the narratives, both assumed to be at position 0. 
    For example, the bar at position +1 indicates that the percentage of the emotion carriers for which the closest filler is the token just after the carrier is almost 6\%, while for random words it is less than 2\%.
    It can be noticed how emotion carriers tend to be more frequently followed by a filler (see positions +1 to +3) within the next 3 words compared to random words.
     }
\label{fig:fillers-carriers}
\end{figure}


\paragraph{Is there a positional relation between emotion carriers and disfluencies?}
To investigate whether there is any recognizable pattern that could be useful for emotion carriers recognition, we analyze emotion carriers annotations in terms of position within narratives. However, we find that emotion carriers annotations are scattered across all positions in the narratives. The mean position of the annotations is usually near the middle of the narratives. \\
Hence, to further deepen the possible positional patterns of distribution of emotion carriers that might be helpful in signaling their occurrence, we focus on disfluencies. 
This intuition was inspired by the results obtained by \newcite{Tammewar2019} on the USoMS corpus.
\newcite{Tammewar2019} show that fillers, a type of disfluency that identifies non-lexical utterances (such as \textit{uhm} or \textit{uh} in English) filling a pause in an utterance or conversation are useful for the task of valence prediction, since they appear both as relevant features in an SVM-based valence prediction approach and analyzing higher attention weights in a Attention-based neural sequence tagger. \\
Inspired by this observation, we study the relative positions of fillers with respect to the emotion carriers to verify whether there is any pattern of association between these elements. For this analysis, we select the nearest fillers for all carriers and summarize the analysis in Figure \ref{fig:fillers-carriers}. Since we are interested in investigating whether fillers occur near the emotion carriers, in the figure we show only the positions $-5$ to $+5$. We find the most occurrences at position $+2$ (7\% of the carriers) followed by $+1$, $+3$ while the most common on the negative side (occur before the carriers) include $-1$, $-4$. To ensure that the proximity of the carriers to fillers is meaningful, we also perform the same analysis for selecting other random words from the same narratives. In order to have a fair comparison, we exclude functional words from the random words considered and select only content words such nouns, adjectives, verbs, adverbs.
Moreover, we only consider as fillers elements that are unambiguously disfluencies in German (i.e. \textit{ähm, äh, mhm}) and do not consider other words (like \textit{also} in German), which depending on the context, may or may not carry meaning. Table \ref{fig:fillers-carriers} shows how, compared to random words, fillers seem to appear more frequently within the immediate context of carriers. While our analysis could be further extended to include other types of disfluencies, this result shows an interesting pattern of association in terms of position between emotion carriers and speech fluency. 

\section{Conclusion and Future Directions}
\label{sect:conclusions}
We proposed a new annotation scheme for the task of extracting emotion carriers from personal narratives (PN), which provides a deeper emotion analysis compared to the conventional emotional prediction task. We performed manual annotations of personal narratives to extract the emotion carriers that best explain the emotional state of the narrator. The annotation was done by four annotators. Narratives being longer and having a complex structure, we find the task to be subjective, which is reflected in the inter-annotator agreement scores and other analyses. Nevertheless, we find surprisingly high overlaps over the annotations, consisting of content words.

As it is difficult to interpret the agreement-scores in itself, to judge the quality of the annotations, we plan to study the improvements in the end-applications making use of the annotations, such as the task of valence prediction. If using the annotations provides better results than the existing systems, we can conclude that the annotations are indeed useful.


We believe that automated extraction of the emotion carriers as a task of Automatic Narrative Understanding (ANU) could benefit various applications. For instance, a conversational agent could use this information (from a PN shared by user) to start a meaningful conversation with the user about the extracted emotion carriers rather than just showing \textit{sympathy} or \textit{happiness} based on the emotion classification of the PN. We plan to annotate the training data and build a module for the automatic extraction of emotion carriers.

Another interesting aspect to study is the correlation between speech and emotion carriers. As our analysis points out, there seems to be an association in terms of position between fillers and carriers, which might be in line with recent findings regarding fillers position and the prediction of perceived metacognitive states \cite{dinkar2020howconfident}.
The USoMs corpus contains speech and transcriptions of the PNs, allowing us to explore the correlation between the different modalities. 
\section{Acknowledgements}
The research leading to these results has received funding from the European Union – H2020 Programme  under grant agreement 826266: COADAPT.
\section{Bibliographical References}
\label{main:ref}
\balance
\bibliographystyle{lrec}
\bibliography{lrec2020W-xample}


\end{document}